\documentclass[runningheads]{llncs}
\usepackage[T1]{fontenc}
\usepackage{graphicx}
\usepackage[ruled]{algorithm2e}
\usepackage{algpseudocode}
\usepackage{subfigure}

\usepackage{amsmath,amssymb,amsfonts}
\usepackage{epstopdf}
\usepackage{textcomp}
\usepackage{xcolor}
\usepackage[ruled]{algorithm2e}
\usepackage{algpseudocode}
\usepackage{svg}
\usepackage{stfloats}
\usepackage{appendix}
\usepackage{marvosym}
\usepackage{wrapfig}
\usepackage{lipsum}
\usepackage{float}

\begin{document}
\bibliographystyle{splncs04}
\title{Wasserstein Diversity-Enriched Regularizer for Hierarchical Reinforcement Learning}
\titlerunning{Wasserstein Diversity-Enriched Regularizer for HRL}
\author{Haorui Li\inst{1,2} \and
Jiaqi Liang\inst{1}\textsuperscript{(\Letter)} \and
Linjing Li\inst{1,2} \and
Daniel Zeng\inst{1,2}}
\authorrunning{H. Li et al.}
%
\institute{State Key Laboratory of Multimodal Artificial Intelligence Systems \\
Institute of Automation, Chinese Academy of Sciences, Beijing, China \and
School of Artificial Intelligence\\
University of Chinese Academy of Sciences, Beijing, China\\
\email{\{lihaorui2021, liangjiaqi2014, linjing.li, dajun.zeng\}@ia.ac.cn}
}
\maketitle              
\begin{abstract}
Hierarchical reinforcement learning composites subpolicies in different hierarchies to accomplish complex tasks.
Automated subpolicies discovery, which does not depend on domain knowledge, is a promising approach to generating subpolicies.
However, the degradation problem is a challenge that existing methods can hardly deal with 
due to the lack of consideration of diversity or the employment of weak regularizers. 
In this paper, we propose a novel task-agnostic regularizer called the Wasserstein Diversity-Enriched Regularizer (WDER), which enlarges the diversity of subpolicies by maximizing the Wasserstein distances among action distributions. 
The proposed WDER can be easily incorporated into the loss function of existing methods to boost their performance further.
Experimental results demonstrate that our WDER improves performance and sample efficiency in comparison with prior work without modifying hyperparameters, which indicates the applicability and robustness of the WDER.

\end{abstract}

\section{Introduction}
Hierarchical reinforcement learning (HRL) decomposes the tasks to be addressed into distinct subtasks and organizes them in a hierarchical structure, 
where the high-level policies solve complex tasks by recombining the low-level subpolicies. 
Through this way, the skills, knowledge, or experience learned by HRL can be shared and reused among different tasks~\cite{r1}.
The transferable ability makes HRL an effective approach to dealing with complex and sparse tasks,
such as multi-level decision-making and fine-grained control over long-horizon manipulation~\cite{r1,r2}, which have made notable progress in recent years.

The generation of subpolicies is the most crucial part of HRL since the quality and diversity of the subpolicies directly affect the performance of the combined policy. Subpolicies can be established by domain experts or be learned automatically. 
Human-designed subpolicies are highly dependent on domain-specific knowledge and meticulously crafted auxiliary pseudo-rewards~\cite{r11}, 
as a result, it is difficult to generalize the obtained subpolicies to new tasks.
By contrast, automated subpolicies discovery aims to learn subpolicies based on simulations with limited input data. 
The automated approach is more demanded as a learning framework that could be applied to various tasks after a little work of adaption. However, it suffers greatly from the degradation problem that all subpolicies degenerate to a common subpolicy in the later stage of the training phase.
The cause of degradation can be imputed to the lack of explicit constraints on the diversity of subpolicies. 
Thus, regularization and rewards reshaping have been employed to mitigate the degradation problem by incorporating information-theoretic objectives, such as maximizing mutual information (MI) and Jensen-Shannon (JS) divergence. 
Nevertheless, the maximum diversity is restricted since both MI and JS are bounded from above and fail to provide an effective gradient when the distributions are supported on non-overlapping domains.
As a result, the degradation problem is still a challenge for training HRL agents.

This paper proposes a Wasserstein Diversity-Enriched Regularizer (WDER) to increase the diversity of subpolicies learned in HRL, which differs from those methods based on the information-theoretic objectives mentioned above. Wasserstein distance (WD) can accurately measure the distribution distance~\cite{r4}, and it provides a geometry-aware topology than traditional $f$-divergences (such as those based on KL divergence). By incorporating a WD-based regularization term in the loss function, the ``distance'' between sequenced subpolicies can be enlarged as far as possible, 
which not only promotes the diversity of the learned subpolicies but also enhances the exploration ability of the composite policy. 
The main contributions of this paper are four folds:
\begin{itemize}
\item We propose a task-agnostic regularizer utilizing WD to enhance the diversity of the learned subpolicies.
\item The proposed regularizer can be easily integrated into various existing HRL methods with a fixed number of subpolicies.
\item We propose a method that applies WDER to two different frameworks: meta-reinforcement learning (Meta-RL)~\cite{r6} and the option framework~\cite{r7}.
\item We evaluate the effectiveness of WDER through two HRL tasks in both discrete and continuous action spaces. The experimental results indicate that our approach outperforms information-theoretic-based methods.
\end{itemize}

\section{Related Works}
This paper is closely related to automated subpolicies generation and WD; thus, we review related work on both in this section.
The formal definition of and how to estimate WD are also introduced to facilitate the expression of our work.

\subsection{Automated Subpolicies Generation}
In RL, the methods for automated subpolicy discovery can be categorized into two families: Unified Learning of Feudal Hierarchy (ULFH) and Unified Learning of Policy Tree (ULPT)~\cite{r10}. In ULFH, a higher-level network called the ``Manager'' samples a subgoal in a learned latent subspace, then a lower-level network called the ``Worker'' must learn a subpolicy to achieve this subgoal~\cite{r12,r13,r45}. Within ULPT, the option framework is a widely applied method that discovers a fixed number of subpolicies in accordance with the learning of a hierarchical policy~\cite{r7,r26}. Meanwhile, Meta Learning Shared Hierarchies (MLSH)~\cite{r6}, an algorithm similar to the ULPT, has been proposed in meta-learning.
MLSH contains a master policy and multiple subpolicies. The master policy employs the same subpolicies in related tasks to accelerate the learning process on unseen tasks. 
Our work is more closely related to enhancing the diversity of subpolicies in ULPT and MLSH, as the main challenge of ULFH is how to design subgoals, 
which is not along the line of ULPT, MLSH, and our work. 

However, the option framework and MLSH training could suffer from the lack of diversity in subpolicies, i.e., different subpolicies converge to nearly the same one. Moreover, in the option framework, the high-level policy may predominantly use only one subpolicy in the entire episode. Some studies have investigated diversity-driven regularizers or reward reshaping through information-theoretic objectives to mitigate this degradation phenomenon. 
Florensa et al.~\cite{r18} introduced a regularizer based on MI between the latent variable and the current state, where the latent variable follows a categorical distribution with uniform weights, in order to increase the diversity of the stochastic neural network policy.  Haarnoja et al.~\cite{r19} obtained diverse policies by maximizing the expected entropy of the trajectory distribution in the reinforcement learning objective. In addition, ``DIAYN''~\cite{r3} forces policies to be diverse and distinguishable by encouraging skills to explore a part of the state space far away from other skills by maximizing entropy in unsupervised RL tasks. 
Huo et al.~\cite{r5} proposed a method using direct JS divergence regularization on the action distributions and emphasized the connection between the visited environment states of subpolicies.

\subsection{Wasserstein Distance}\label{section2b}
The measurement of discrepancy or distance between two probability distributions can be treated as a transport problem~\cite{r23}. 
Let $p$ be a probability distribution defined on domain $\mathcal{X}\subseteq\mathbb{R}^{n}$ and $q$ be a distribution defined on $\mathcal{Y}\subseteq\mathbb{R}^{m}$. Let $\mathrm{\Gamma}[p, q]$ be the set of all distributions on the product space $\mathcal{X\times Y}$, with their marginal distributions on $\mathcal{X}$ and $\mathcal{Y}$ being $p$ and $q$, respectively. Thus, given an appropriate cost function $c(x, y) :\mathcal{X\times Y\rightarrow \mathbb{R}}$ which represents the cost of moving a unit ``mass'' from $x$ to $y$, the WD is defined as
\begin{equation}
W_{c}(p, q)=\inf _{\gamma \in \mathrm{\Gamma}[p, q]} \int_{\mathcal{X} \times \mathcal{Y}} c(x, y) d \gamma. \label{eq1}
\end{equation}

The optimal transport is the one that minimizes the above transport cost. 
The smoothed WD is introduced to address the challenge of super-cubic complexity:
\begin{equation}
\widetilde{W}_{c}(p, q)=\inf _{\gamma \in \mathrm{\Gamma}[p, q]}\left[\int_{\mathcal{X} \times \mathcal{Y}} c(x, y) d \gamma+\beta K L(\gamma \mid p,q)\right].
\label{eq2}
\end{equation}

Eq.~\eqref{eq2} can be estimated either by the primal or dual formulation. In this paper, we calculate the WD by the dual formulation. 
The dual formulation is based on the Fenchel-Rockafellar duality~\cite{r23}, which provides a convenient neural way to estimate WD. 
Let set $\mathcal{A}=\{(u, v) \mid \forall(x, y) \in \mathcal{X} \times \mathcal{Y}: u(x)-v(y) \leq   c(x, y)\} $, 
where $\mu: \mathcal{X} \rightarrow \mathbb{R}$ and $\nu: \mathcal{Y} \rightarrow \mathbb{R}$ are continuous functions 
and the cost function $c(x, y)$ may not be smooth, then the dual formulation estimation of WD is
\begin{equation}
W_{c}(p, q)=\sup _{(\mu, \nu) \in \mathcal{A}} \underset{x \sim p(x), y \sim q(y)}{\mathbb{E}}[\mu(x)-\nu(y)].
\label{eq3}
\end{equation}
Theoretically, the maximum value obtained from the dual formulation aligns with the minimum value of the original formulation.
Accordingly, the dual formulation of the smoothed WD is 
\begin{equation}
\widetilde{W}_{c}(p, q)=  \sup _{\mu, \nu} \underset{x \sim p(x), y \sim q(y)}{\mathbb{E}}\left[\mu(x)-\nu(y)
 -\beta \exp \left(\frac{\mu(x)-\nu(y)-c(x, y)}{\beta}\right)\right].
\label{eq4}
\end{equation}
The dual formulation is more convenient, as it does not impose any constraints on the functions $\mu$ and $\nu$.

WD has been widely used to quantify distribution differences in representation learning~\cite{r27} and generative modeling~\cite{r29}. 
In the context of RL, WD has been used to quantify the difference between policies~\cite{r28}.
Compared to traditional KL and other $f$-divergences, WD has shown to be a versatile measure. 
Pacchiano et al.~\cite{r22} adopted WD to enhance the performance of trust region policy optimization and evolutionary strategies, 
Dadashi et al.~\cite{r31} showed its effectiveness in imitation learning by minimizing it between the state-action distributions of the expert and the agent, Moskovitz et al.~\cite{r39} used it as a divergence penalty with the local geometry to speed optimization. 
Furthermore, WD was employed as a metric for unsupervised RL to encourage the agent to explore the state space extensively to generate diverse subpolicies~\cite{r4}.
Different from these studies, our WDER utilizes an action distributions-based regularization term, WD is employed to measure the differences in the learning phase of different subpolicies.

\begin{figure}[tb]
\centering
\includegraphics[height=6cm]{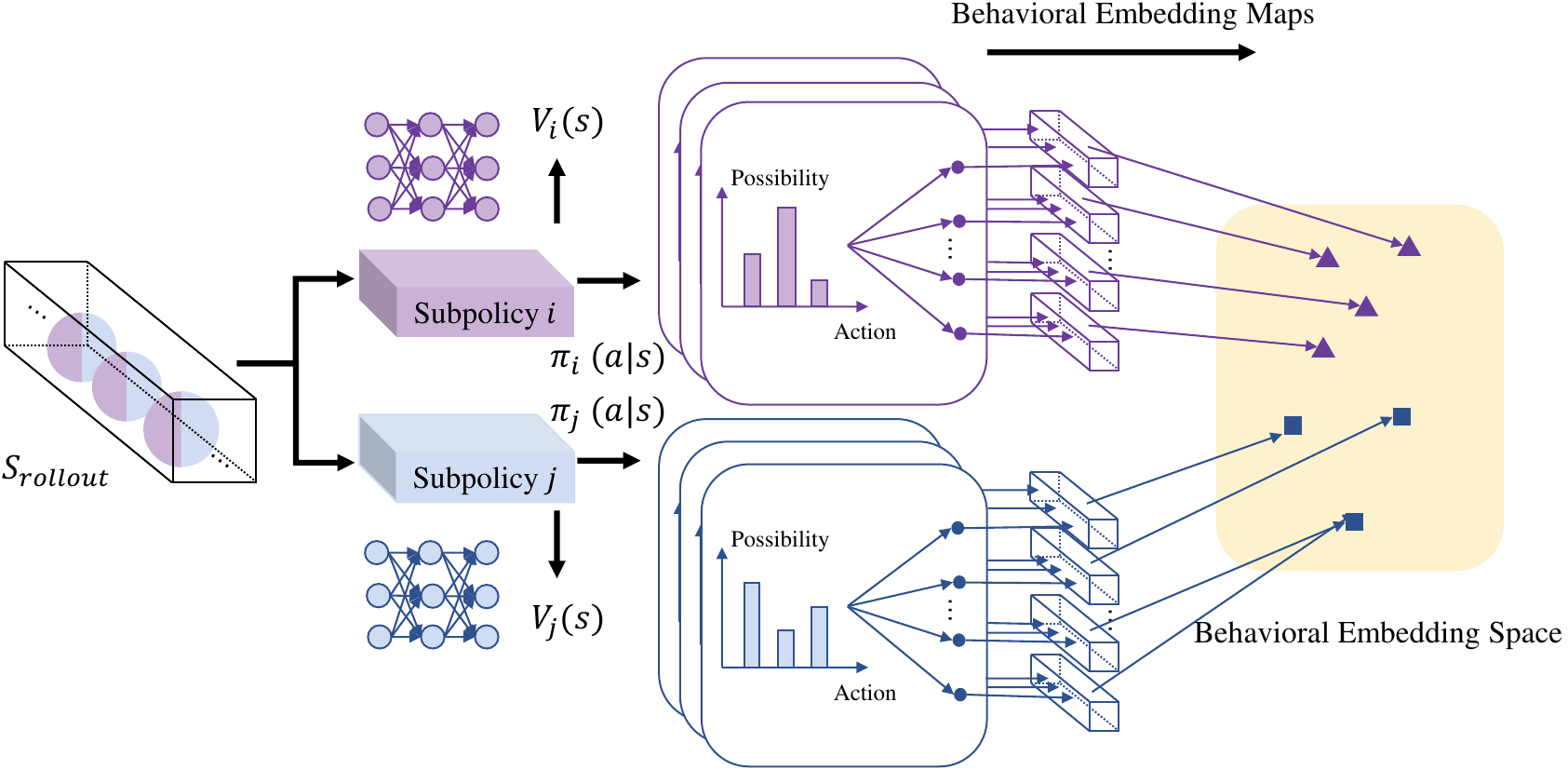}   
\caption{Estimation of the WD between two subpolicies $\pi_i$ and $\pi_j$. 
The actions generated by subpolicy $\pi_i$ and $\pi_j$ at the same states in the state set $S_{rollout}$ are sampled and mapped to feature vectors in the behavior embedding space 
according to the behavioral embedding maps. 
Two sampled action vectors can be mapped to the same point in the behavior embedding space. After the mapping is completed, we can estimate the WD by utilizing Alg.~\ref{algorithm1} and Eq.~\eqref{eq5}.}
\label{figure0}
\end{figure}

\section{Methodology}
\subsection{Standard RL and HRL}
This paper adopts the standard RL setting, which is built upon the theory of Markov decision processes (MDPs)~\cite{r21}. 
A MDP can be formalized as a tuple $\langle\mathcal{S}, \mathcal{A}, \mathcal{R}, \mathcal{P}, \mathcal{\gamma}\rangle$, 
where $\mathcal{S}$ is a finite set of states, $\mathcal{A}$ is a finite set of actions, $\mathcal{R: S \times A \rightarrow} [R_{min}, R_{max}]$ is the reward function, $\mathcal{P}: \mathcal{S} \times \mathcal{A} \rightarrow \Delta(\mathcal{S})$ is the transition function, 
where $\Delta (\mathcal{S})$ denotes the set of probability distributions over $\mathcal{S}$, and $\gamma\in[0,1)$ is the horizon discount factor. 
This paper focuses on a standard RL agent with a two-level architecture, and there are $\mathit{K}$ low-level subpolicies denoted as $\pi _{1}, \pi _{2}, \cdots,\pi _{K}$,
and a high-level master policy $\pi _{m}$ that decides which subpolicy to be used in the current state. 
At each time step $t$, the currently selected subpolicy $\pi_{k}$ samples an action $a_{t}$ based on the observed state $s_{t}$ with respect to the distribution $\pi_{k} \left ( s_{t} \right )$. The execution of the action $a_{t}$ results in the generation of an environmental reward $r_{t}$ and the transition of the system to a new state $s_{t+1}$ according to the transition probability $\mathit{p}\left ( s_{t+1}\mid s_{t}, a_{t} \right )$. This process continues iteratively until the master policy $\pi _{m}$ selects another subpolicy.

\subsection{Wasserstein Distance between Subpolicies}\label{section3b}

To estimate the WD between two subpolicies, we need to obtain the policy embeddings~\cite{r22} of these two subpolicies. Fig.\ref{figure0} depicts the process involved. In the context of HRL, each subpolicy $\pi_{k}$ can be fully defined by its action probability distribution. 
In settings with discrete action spaces, the action probability distribution of a subpolicy typically follows a categorical distribution. 
While in the case of continuous action spaces, the action probability distribution of a subpolicy commonly follows a Gaussian distribution~\cite{r33}, which means that the agent samples an action $a_{t} \sim \mathcal{N}\left ( \mu,\sigma   \right )$ at each time step.
The mean $\mu$ and the standard deviation $\sigma$ are obtained by fitting a normal distribution according to the outputs of the last dense layer of $\pi_{k}$'s policy network.

 

\begin{algorithm}[htbp]
\caption{Random Features Wasserstein SGD}\label{algorithm1}
\KwIn{kernels $\kappa$, $\ell$ over $\mathcal{X}$, $\mathcal{Y}$ respectively with corresponding random feature maps $\phi_{\kappa}$, $\phi_{\ell}$, smoothing parameter $\gamma$, gradient step size $\eta$, number of optimization rounds $M$, initial dual vectors $\mathbf{p}_{0}^{\mu}$, $\mathbf{p}_{0}^{\nu}$.}
\For {$t=0, \cdots, M$}
{Sample $\left(x_{t}, y_{t}\right) \sim \mu \otimes \nu$.

Update $\mathbf{p}_t^\mu, ~\mathbf{p}_t^\nu$ using Eq.~\eqref{update}.



}
\KwOut{$\mathbf{p}_{M}^{\mu}$, $\mathbf{p}_{M}^{\nu}$}
\end{algorithm}

Next, we elucidate the sampling procedure, outlined in Alg.~\ref{algorithm1}~\cite{r22}, when estimating the WD between subpolicy $\pi_k$ and $\pi_l$.
First, we extract a set of $T$ states, denoted as $S_{rollout}$, from the trajectories generated by these two subpolicies. The states within $S_{rollout}$ are represented as $s_1, s_2, \cdots, s_T$. For each state $s_t$, we then sample $B$ actions from the action probability distributions generated by the policy networks of $\pi_k$ and $\pi_l$. These sampled actions are denoted as $\left\{a_{kti}\right\}_{i=1}^{B}$ and $\left\{a_{lti}\right\}_{i=1}^{B}$, respectively. To mitigate the variance of the cost function $c(x, y)$ in Eq.~\eqref{eq4}, we employ the same sequence of random numbers, i.e., the common random numbers~\cite{r32}, during the generation of both $\left\{a_{kti}\right\}_{i=1}^{B}$ and $\left\{a_{lti}\right\}_{i=1}^{B}$.

Second, we map the sampled actions to the embedding space by a radial basis function (RBF) kernel using random Fourier feature maps~\cite{r35}. 
This process is referred to as the behavioral embedding map (BEM) :
\begin{equation}
\mathrm{\Phi}: \mathrm{\Gamma}[p_k, q_l] \rightarrow \mathcal{E}, 
\end{equation}
where $\mathrm{\Gamma}[p_k, q_l]$ bears the same meaning as $\mathrm{\Gamma}[p, q]$ defined in Section~\ref{section2b}. 
In this context, the notations $p_k$ and $q_l$ denote the action probability distributions corresponding to subpolicies $\pi_k$ and $\pi_l$. 
While $\mathcal{E}$ corresponds to the embedding space and can be regarded as a behavioral manifold~\cite{r22}, 
wherein each action (of dimension $m$) is meticulously mapped into a vector of features (of dimension $d > m$). 
The BEM $\mathrm{\Phi}$ induces a corresponding pushforward distribution, i.e., the resulting distribution, on $\mathcal{E}$.
For subpolicies $\pi_{k}$ and $\pi_{l}$, we denote their pushforward distributions as $\mathbb{P}_{\pi_{k}}^{\mathrm{\Phi}}$ and $\mathbb{P}_{\pi_{l}}^{\mathrm{\Phi}}$, respectively.

For subpolicies $\pi_{k}$ and $\pi_{l}$, we define $\mu(x)$ and $\nu(y)$ in Eq.~\eqref{eq4} as
\begin{equation}
    \mu(x) = \left(\mathbf{p}^{\mu}\right)^{\top}\phi(x),~~~~ \nu(y)=\left(\mathbf{p}^{\nu}\right)^{\top}\phi(y),
\end{equation}
where $\mathbf{p}^{\mu},\ \mathbf{p}^{\nu} \in \mathbb{R}^{m}$ are vectors with $m$ random features, $\phi(x)$ is defined as $\phi(x) = \frac{1}{\sqrt{m}} \cos (x\mathbf{G}+\mathbf{b})$, $x$ and $y$ belong to $\mathbb{R}^{h\times d}$, 
$h=B \times T_{\text{minibatch}}$ represents the number of actions in a minibatch, 
$d$ denotes the dimensionality of the action space. $\mathbf{G} \in \mathbb{R}^{d \times m}$ is a Gaussian matrix with iid entries sampled from $\mathcal{N}(0,1)$. The vector $\mathbf{b} \in \mathbb{R}^{m}$ is composed of independently sampled elements from the uniform distribution $\operatorname{U}[0,2\pi]$, and the $\cos(\cdot)$ function is applied elementwise. 

Then we can find the optimal dual estimation of WD by Alg.~\ref{algorithm1} employing Random Features Wasserstein Stochastic Gradient Descent (SGD). Given the input kernels $\kappa$, $\ell$, and a fresh sample $\left(x_{t}, y_{t}\right) \sim \mu \otimes \nu$, where $\otimes$ represents the tensor product, the parameters w.r.t. the current iteration should satisfy:
\begin{equation}
\begin{aligned}
F\left(\mathbf{p}_{t}^{\mu}, \mathbf{p}_{t}^{\nu}, x_{t}, y_{t}\right) & = 
\exp \left(\frac{\left(\mathbf{p}_{t}^{\mu}\right)^{\top} \phi_{\kappa}(x_{t})-\left(\mathbf{p}_{t}^{\nu}\right)^{\top} \phi_{\ell}(y_{t})-C(x_{t}, y_{t})}{\gamma}\right),
\\
\left(\begin{array}{l}
\mathbf{p}_{t+1}^\mu \\
\mathbf{p}_{t+1}^\nu
\end{array}\right) & =\left(\begin{array}{l}
\mathbf{p}_t^\mu \\
\mathbf{p}_t^\nu
\end{array}\right)+\left(1-F\left(\mathbf{p}_t^\mu, \mathbf{p}_t^\nu, x_t, y_t\right)\right) v_t,
\end{aligned}\label{update}
\end{equation}
where $v_{t}={\eta}\left(\phi_{\kappa}\left(x_{t}\right)-\phi_{\ell}\left(y_{t}\right)\right)^{\top}$. Let $M$ be the maximum number of iteration, $\mathbf{p}^{\mu}_M$ and $\mathbf{p}^{\nu}_M$ be the output of Alg.\ref{algorithm1}.
We can estimate the WD between two subpolicies $\pi_k$ and $\pi_l$ using $\mathbf{p}^{\mu}_M$ and $\mathbf{p}^{\nu}_M$ as


\begin{equation}
\mathrm{WD}_{\gamma}\left(\mathbb{P}_{\pi_{k}}^{\mathrm{\Phi}}, \mathbb{P}_{\pi_{l}}^{\mathrm{\Phi}}\right) =  \hat{\mathbb{E}}\left[\left(\mathbf{p}_{M}^{\mu}\right)^{\top} \phi_{\kappa}(x_{i}) - \left(\mathbf{p}_{M}^{\nu}\right)^{\top} \phi_{\ell}(y_{i}) - \frac{F\left(\mathbf{p}_{M}^{\mu}, \mathbf{p}_{M}^{\nu}, x_{i}, y_{i}\right)}{\gamma}\right],
\label{eq5}
\end{equation}
where $\hat{\mathbb{E}}$ denotes the empirical expectation over $C$ iid action samples $\left\{\left(x_{i}, y_{i}\right)\right\}_{i=1}^{C}$,
$x_{i}$ and $y_{i}$ correspond to the sampled actions of subpolicies $\pi_{k}$ and $\pi_{l}$, respectively. 

\subsection{HRL with Wasserstein Diversity-Enriched Regularizer}
\begin{wrapfigure}{htbp}{0.45\textwidth}  
  \begin{minipage}{\linewidth}
    \centering
    \includegraphics[width=\linewidth]{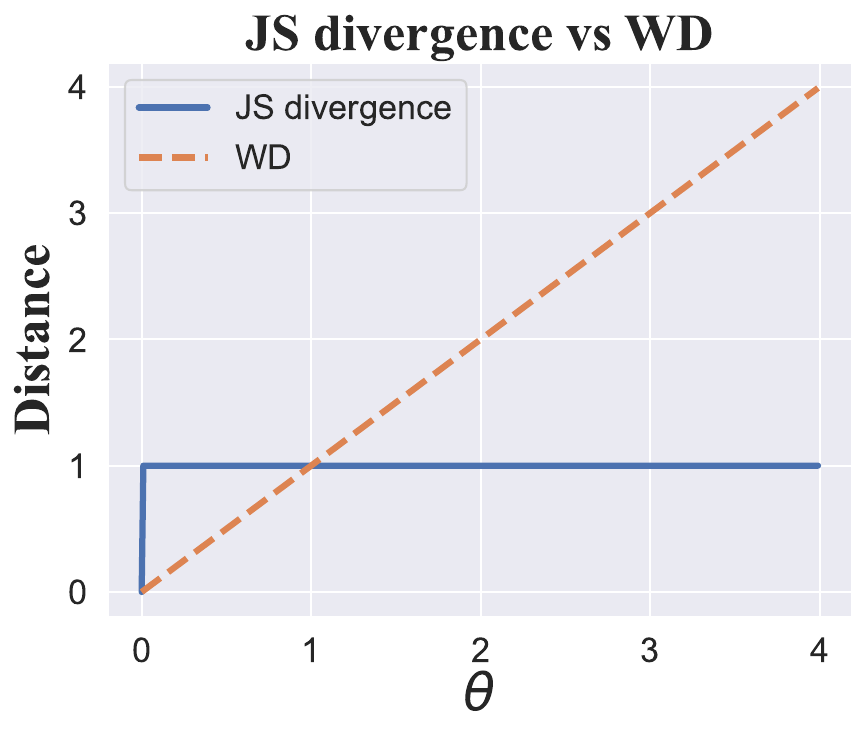}  
    \caption{Examples of JS divergence and WD between distributions $P$ and $Q$, where $\forall(x, y) \in P, ~ x=0 \text { and }y \sim U(0,1)$ and $\forall(x, y) \in Q, ~x=\theta, \theta \geq 0 \text { and }y \sim U(0,1)$. The WD provides useful distance information between $P$ and $Q$, while the JS divergence fails.}
    \label{JS_and_WD}
  \end{minipage}
\end{wrapfigure}
Regularization is an effective and convenient framework to help generate diverse subpolicies,
the maximum value of the regularizer and the distributions taken into consideration by the regularization are the most important determinants.
Information-theoretic measures, such as MI and JS divergence, are the commonly encountered regularization form~\cite{r3,r18,r19}.
However, these measures are bounded from above, which intrinsically limits the diversity that can be achieved.
Taking the classical MI as an illustration, the MI between two random variables $S$ and $Z$ is $I(S; Z)=H(Z)-H(Z|S)$, which is bounded by $H(Z)$,
where $H(\cdot)$ is the Shannon entropy. The MI reaches its maximum value when $H(Z|S)=0$, i.e., the support of $S$ and $Z$ do not overlap.
The JS divergence has the same limitation when adopted to encourage diverse subpolicies.
With this upper bound, the distance between two distributions cannot be enlarged; even their essential difference can still be amplified.
Different from information-theoretic measures, the upper bound of WD can be set to the predefined value by choosing the appropriate cost function $c(x, y)$ according to Eq.~(\ref{eq1}). This is a highly demanded property to encourage diverse subpolicies in HRL~\cite{r4}. 
In addition, WD can provide smooth and informative gradients for updating parameters, regardless of whether the distributions of the two subpolicies overlap or not, while MI and JS cannot.
Fig.~\ref{JS_and_WD} illustrates the changes of the JS divergence and WD between distributions $P$ and $Q$ with respect to $\theta $. When the distributions do not overlap, the WD still provides useful information about the distance between the distributions, while the JS divergence does not.

As to the distributions, few works directly took the action distributions as the inputs~\cite{r5,r26}. Differentiating strategies based on action probability distribution has advantages that are not exhibited by state probability distribution-based strategies. It helps the agent explore more action choices, leading to the discovery of better subpolicies. It also allows the agent to adjust its action choices according to different situations, which improves its ability to adapt to various environments and states. Moreover, this approach enhances the agent's capability to handle complex policy spaces by providing it with a rich range of policy expressions.
To sum up, in this paper, we devise the WDER and take the action distributions as the input for automatically generating highly diverse subpolicies in HRL.

Given $N\ge 2$ subpolicies, for an arbitrarily chosen subpolicy $\pi_k$, to make it distinct from other subpolicies, we want to maintain a distance (the larger, the better) of $\pi_k$ with respect to others. Based on WD, We adopt the regularizer as
\begin{equation}
\mathrm{WD}_{min}(\pi_{k}) = \min _{j \neq k}\mathrm{WD}_{\gamma}\left(\mathbb{P}_{\pi_{k}}^{\mathrm{\Phi}}, \mathbb{P}_{\pi_{j}}^{\mathrm{\Phi}}\right),
\label{wd_min}
\end{equation}
 to make $\pi_{k}$ away from its nearest subpolicy. By WDER, the actor networks are encouraged to converge in different local maxima~\cite{r22}.
For subpolicy $\pi_k$, let $\theta_{\pi k}$ and ${\theta _{vk}}$ be the parameters of the policy network and the value network in the actor-critic framework, respectively. The modified actor network incorporates $\mathrm{WD}_{min}(\pi_{k})$ as the regularization term in its loss function
\begin{equation}
L_{new}\left(\theta_{\pi k}\right) = L_{old}\left(\theta_{\pi k}\right) -\alpha \mathrm{WD}_{min}(\pi_{k}), 
\label{loss_with_wder}
\end{equation}
where $L_{old}\left(\theta_{\pi k}\right)$ is depended on the baseline and $\alpha$ is a hyperparameter. 

The choice of backend training algorithm for RL can be different according to the nature and settings of the specific problems. As the PPO (Proximal Policy Optimization)~\cite{r42} is adaptable to both discrete and continuous action space, we employ it as the backend RL algorithm. Specifically, Alg.~\ref{algorithm2} outlines the proposed method, where HRL$_\mathrm{base}$ denotes the input baseline. The algorithm outputs the trained model with the parameters of the subpolicies described by $\theta_{\pi k}$ and ${\theta _{vk}}$, $k=1,2,\cdots, K$, 
together with the parameters of the master policy $\pi_m$ updated according to the loss function of the baseline master policy.


\begin{algorithm}[htbp]
\caption{Training Algorithm for the WDER Approach}\label{algorithm2}
\KwIn{The baseline method HRL$_\mathrm{base}$, and regularization hyperparameters $\alpha $}
\While{not convergence}{\While{episode not terminates}{Sample a subpolicy $\pi _{k}$ from the master policy $\pi _{m}$ according to HRL$_\mathrm{base}$\;
\While{$\pi _{k}$ not terminates}{Sample $a _{t}$ according to the subpolicy $\pi _{k}(a_{t} \mid s_{t})$, and collect the critic network output $v_{k}\left (s_{t} \right )$ from $\pi _{k}$\;
Perform action $a _{t}$ and receive reward $r _{t}$ and get the next state from the transition probability $\mathit{p}\left ( s_{t+1}\mid s_{t},a_{t} \right ) $\;
Collect experience $s _{t}$, $a _{t}$, $r _{t}$, $s _{t+1}$, $v_{k}\left (s_{t} \right )$;
}}
update the actor network and the critic network of the master policy $\pi _{m}$ according to the loss function of the baseline master policy

\For {each subpolicy $\pi _{k}$ with parameters $\left \{   \theta_{\pi k}, \theta _{vk} \right \} $}
{update parameter ${\theta_{\pi k}}$ following the Eq.~\eqref{loss_with_wder}


update parameter ${\theta _{vk}}$;\
}}
\KwOut{The trained model by the WDER-augmented HRL$_\mathrm{base}$}
\end{algorithm}

As to the complexity of the proposed method, 
the distance computing time grows in the order of $O(N^{2})$ as the number $N$ of subpolicies increase~\cite{r4}.
When $N$ is large, the WD can be approximated by the sliced or projected WD~\cite{r37,r38,r40}, and some experiments indict that  
HRL algorithm with two subpolicies achieves the best performance in most practical applications~\cite{r26}. 
Based on this result, we use two subpolicies in our experiments. 

\section{Experiment}
We examine our method on two typical RL domains and select the corresponding state-of-the-art approaches for comparison: 1) MLSH for Meta-RL~\cite{r6}; 2) OC (option-critic) for the option framework~\cite{r7}. 

\subsection{Variant Algorithms and Experimental Setup}
\paragraph{WDER-MLSH} The MLSH architecture contains a master policy and several subpolicies. Our proposed WDER method can be easily integrated into the loss functions of the MLSH subpolicies. We compare the generalization performance of our WDER-MLSH with the original MLSH approach on a discrete 2-D navigation task. We also evaluate the ability to adapt to new tasks of our approach with $\alpha$ setting to 0.5. Other hyperparameters are the same as used in~\cite{r6}.
\paragraph{WDER-OC} As a significant component of HRL, the option framework has its own learning and optimization system. To test our WDER approach on high-dimensional input tasks, we evaluate its performance with the original OC on complex robot tasks. For these experiments, we set the value of the corresponding parameter $\alpha$ as 0.2 and use two options. We adopt the hyperparameters and convolution structure settings used in~\cite{r7}.

\subsection{Performance Evaluation}
We first investigate the performance of subpolicy discovery in Meta-RL by evaluating our WDER-MLSH algorithm on MovementBandits, viz a 2-D navigation task~\cite{r6}. 
In this task, an agent is placed in a planar world, and it has already known its current location and the candidate target locations. The agent can take discrete actions to move in four directions or remain stationary. The environment sends 1 to the agent as a reward if the agent is at a certain distance from the correct target point and 0 otherwise. We use two subpolicies to train the MLSH baseline, MLSH-JS with JS divergence regularizer implemented following~\cite{r5} and our WDER-MLSH method, with the duration of each subpolicy being ten timesteps. We vary the coefficient of the WDER term $\alpha$ from 0.2 to 0.6 with step 0.1.

\begin{figure}[tbp]
\centering
\subfigure[]
{
    \begin{minipage}[h]{0.47\linewidth}
        \centering
        \includegraphics[scale=0.33]{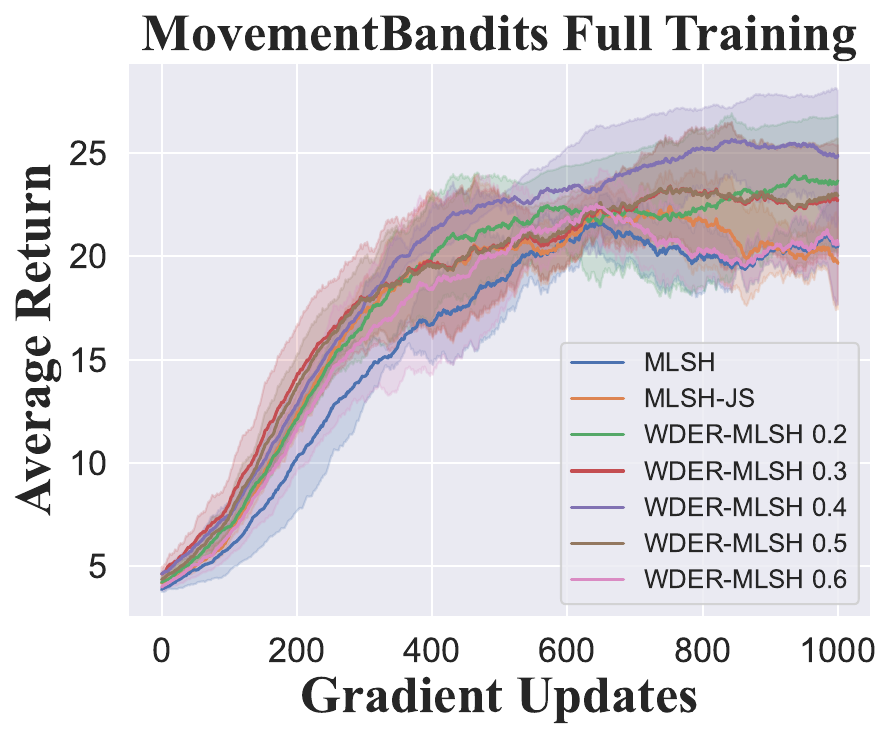}
        \label{figure1a}
    \end{minipage}
}
\subfigure[]
{
 	\begin{minipage}[h]{.47\linewidth}
        \centering
        \includegraphics[scale=0.33]{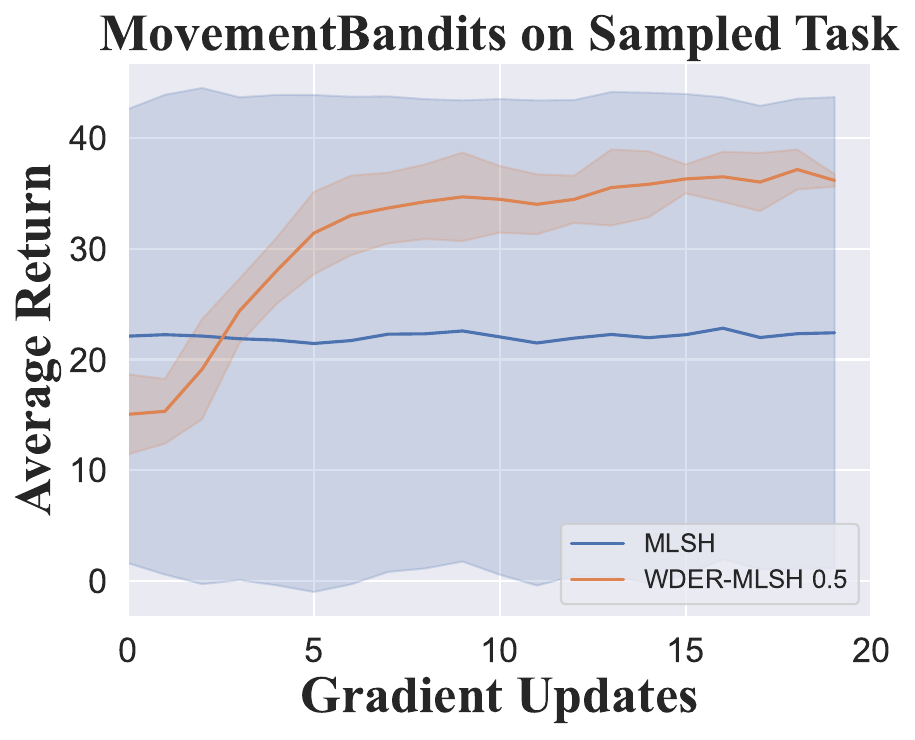}
        \label{figure1b}
    \end{minipage}
}
\caption{(a) Average return curves of WDER-MLSH with different $\alpha $ during the training phase in the MovementBandits environment. (b) Average return curves for the newly sampled MovementBandits task. Performance corresponds to the ability to adapt to new tasks.}
\label{figure1}
\end{figure}

\begin{figure*}[tb]
\centering
\subfigure
{
    \begin{minipage}[b]{.8\linewidth}
        \centering
        \includegraphics[scale=0.4]{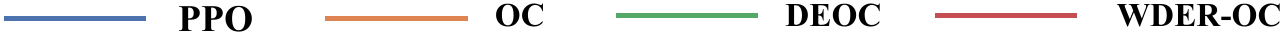}
    \end{minipage}
}
\subfigure
{
    \begin{minipage}[b]{.45\linewidth}
        \centering
        \includegraphics[scale=0.30]{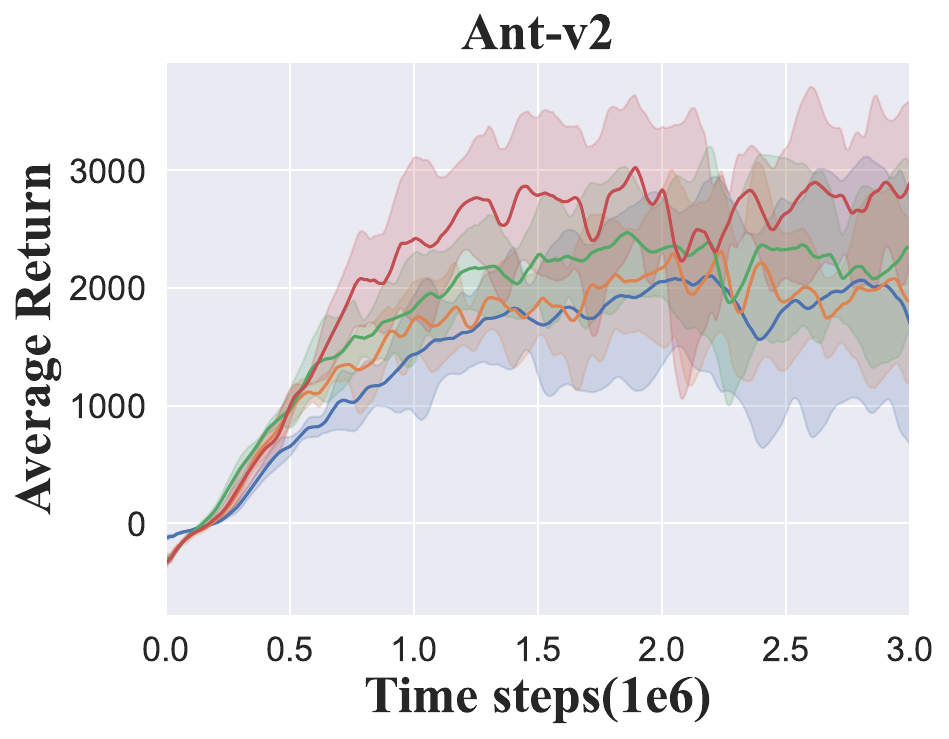}
    \end{minipage}
}
\subfigure
{
 	\begin{minipage}[b]{.45\linewidth}
        \centering
        \includegraphics[scale=0.30]{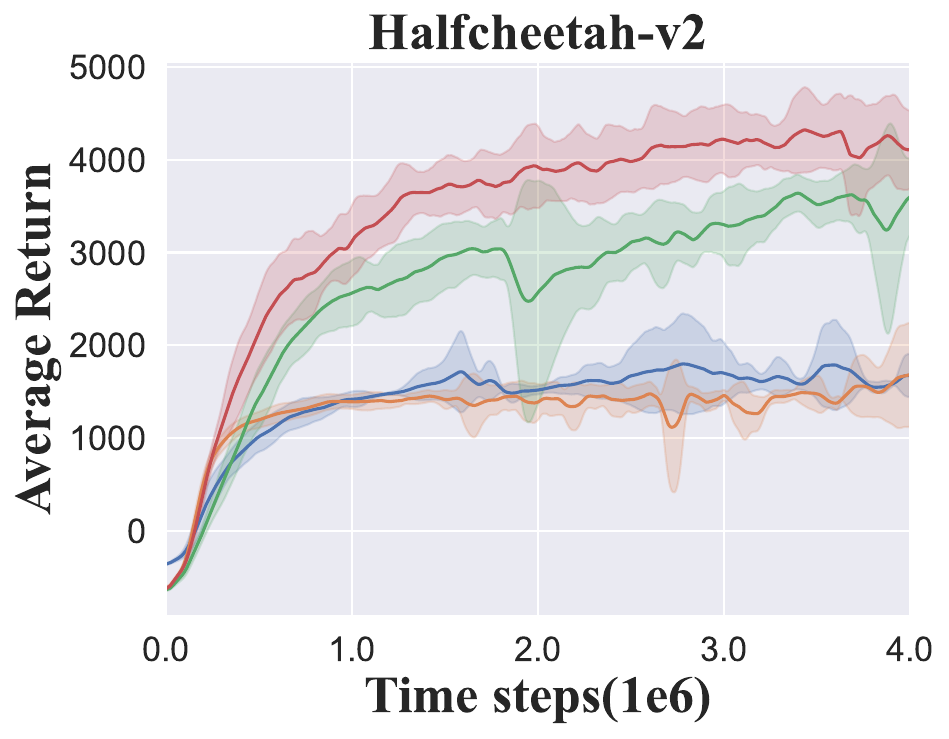}
    \end{minipage}
}
\subfigure
{
 	\begin{minipage}[b]{.45\linewidth}
        \centering
        \includegraphics[scale=0.30]{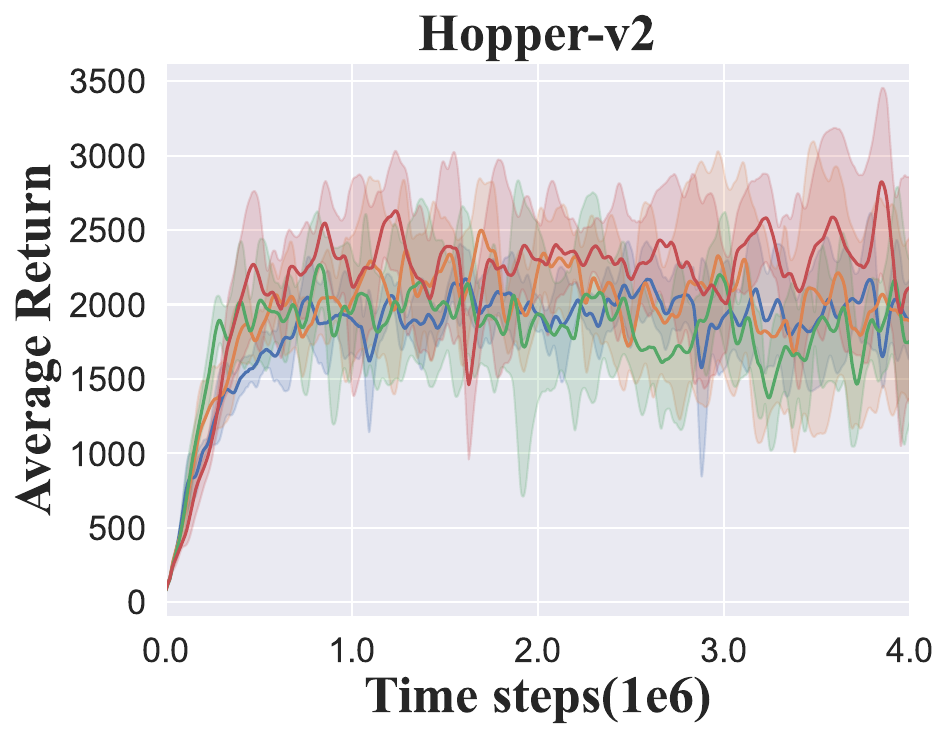}
    \end{minipage}
}
\subfigure
{
 	\begin{minipage}[b]{.45\linewidth}
        \centering
        \includegraphics[scale=0.30]{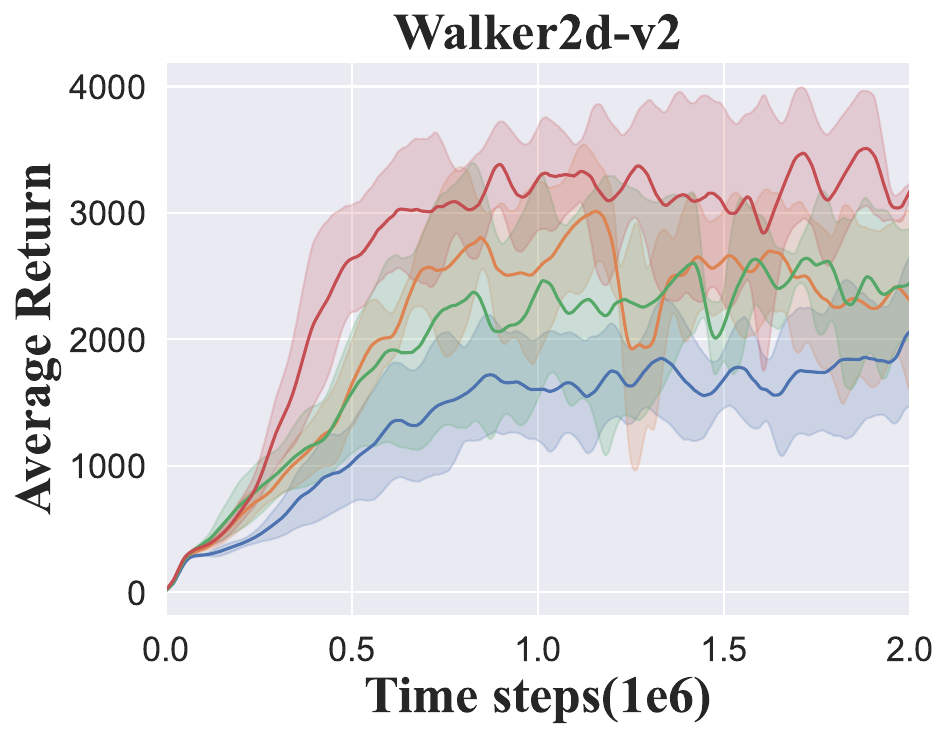}
    \end{minipage}
}
\caption{Average return curves of WDER-OC on standard Mujoco tasks.}
\label{figure3}
\end{figure*}

\begin{figure*}[tb]
\centering
\includegraphics[scale=0.44]{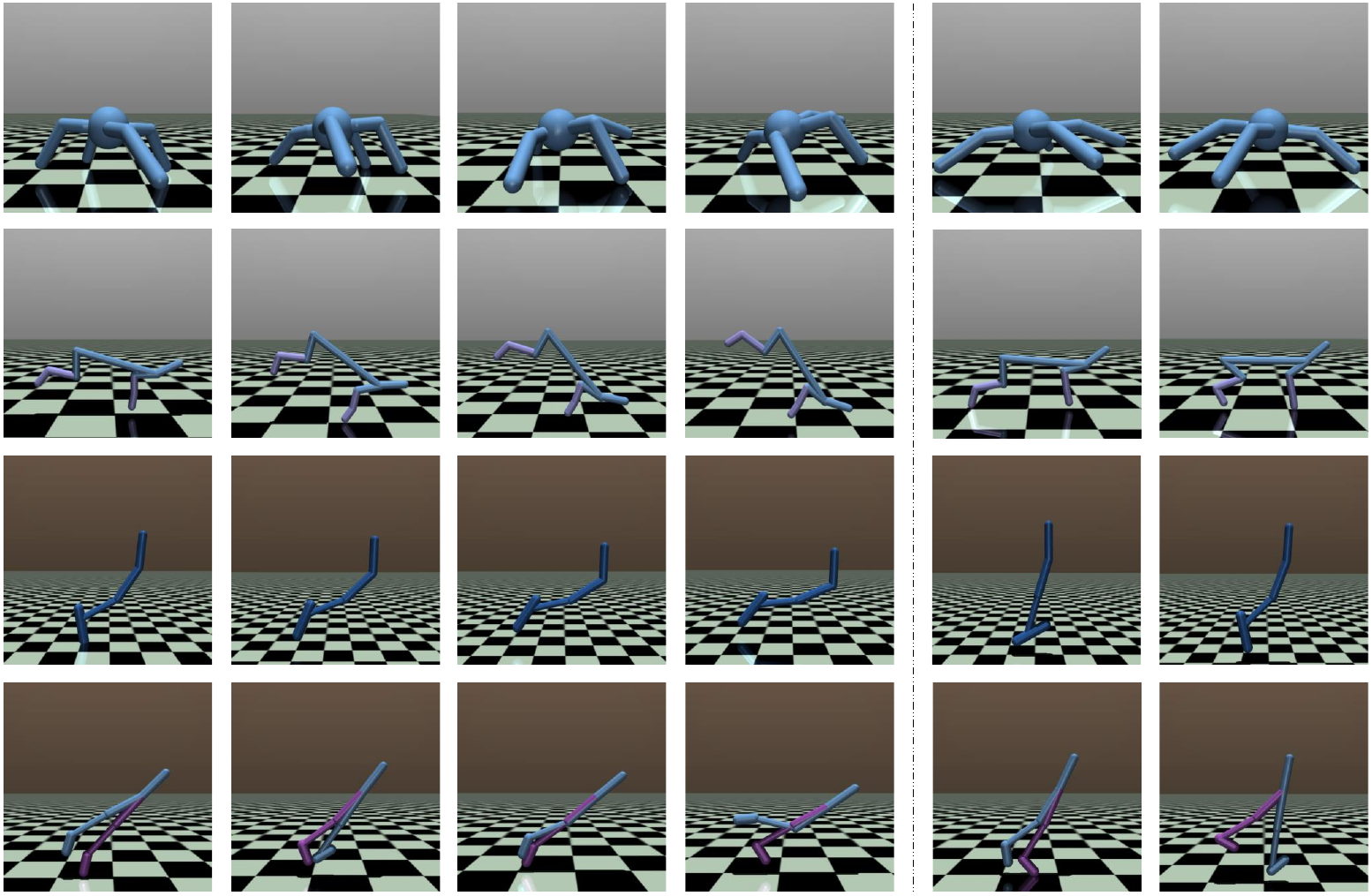}
\caption{Six frames of Ant, Halfcheetah, Hopper, and Walker2d, respectively. The first four images in each row illustrate scenarios of task failures frequently encountered during the execution of the baselines. The last two images in each row showcase the stable performance of our WDER in the exact scenarios. Our WDER approach significantly reduces the occurrence of task failures and achieves superior performance.}
\label{figure5}
\end{figure*}


The average return curves of different agents during the training phase are shown in Fig.~\ref{figure1a}, where each line is averaged over three runs, and the shaded areas represent one standard deviation. Our WDER-MLSH agents outperform the MLSH agent with respect to the average return. Especially when setting $\alpha$ as 0.5, 
the WDER-MLSH agent outperforms the MLSH baseline by approximately 25$\%$ and achieves superior performance using only 30$\%$ of all samples. 
Furthermore, the steeper learning curves of WDER-MLSH in the early stages demonstrate that these agents have remarkable sample efficiency and can quickly improve their overall performance by mastering some fundamental skills.

Further, we assess the transferability of each agent by examining the diversity and effectiveness of the subpolicies learned in new tasks. Specifically, we train the agents in a MovementBandits environment until the cumulative reward stabilizes, then freeze the learned subpolicies, and then fine-tune the high-level strategy. 
We conduct six independent runs to compare the results with the original MLSH. Fig.~\ref{figure1b} indicates that our approach surpasses the baseline by converging faster and achieving superior performance in nearly 20 steps. These results suggest that our method improves Meta-RL performance for new tasks.

\begingroup
\renewcommand{\arraystretch}{1.25} 
\begin{table}[htbp]
\caption{The max average returns on four Mujoco tasks over 4 million timesteps}
\begin{center}
\tabcolsep=0.49cm
\begin{tabular}{c|cccc}
\hline
               & Ant & Halfcheetah  & Hopper  & Walker2d  \\ \hline
PPO         & 2103.2 & 1799.0 & 2172.7 & 2083.1  \\
OC          & 2310.6 & 1682.3 & 2498.6 & 3012.6  \\
DEOC        & 2471.5 & 3640.4 & 2267.2 & 2778.0  \\ 
WDER-OC     & \textbf{3023.9} & \textbf{4322.8} & \textbf{2823.7} & \textbf{3510.1}  \\ \hline
\end{tabular}
\label{table1}
\end{center}
\end{table}
\endgroup

Next, we evaluate the generalization ability of our WDER algorithm in the option framework on four classic Mujoco tasks~\cite{r36}. We compare it with a standard OC approach with two options. Furthermore, as we implement our option-critic method using PPO~\cite{r42}, we also report the results obtained through PPO as a reference. Our comparisons include an information-theoretic intrinsic reward method (diversity-enriched option-critic, DEOC)~\cite{r26}. 
The return curves in the training phase are averaged over five independent runs and smoothed by a sliding window of size 20 (Fig.~\ref{figure3}). Table~\ref{table1} shows the max average returns over $4$ million timesteps. The results indicate that WDER-OC surpasses all three baselines, particularly on the Halfcheetah task, where the performance is 157$\%$ higher than the original OC baseline and with less variance. Hence, our WDER algorithm is shown to be more effective in improving the performance of the option framework.

\begingroup
\renewcommand{\arraystretch}{1.25} 
\begin{table}[!h]
\caption{The average returns on four classical Mujoco tasks over 1 million timesteps}
\begin{center}
\tabcolsep=0.49cm
\begin{tabular}{c|cccc}
\hline
        & Ant    & Halfcheetah & Hopper & Walker2d \\ \hline
DAC+PPO & 985.8  & 1830.1      & 1702.2 & 1968.0   \\
AHP+PPO & 1359.3 & 1701.7      & 1993.6 & 1520.6   \\
MOPG    & 907.4  & \textbf{3446.7} & 1955.3 & 1856.9   \\ 
WDER-OC & \textbf{2418.9} & 3119.5 & \textbf{2128.1} & \textbf{2368.5}   \\ \hline
\end{tabular}
\label{table2}
\end{center}
\end{table}
\endgroup
In order to validate whether WDER-OC can outperform other option variants and non-option baselines, we compared WDER-OC with DAC+PPO~\cite{r43}, AHP+PPO~\cite{r44}, and MOPG~\cite{r41}. Since MOPG uses the least timesteps among these algorithms, we compare the performance of these algorithms with WDER-OC after running for 1 million time steps as shown in Table~\ref{table2}. The results for all algorithms except WDER-OC are the same as reported in the MOPG paper.
It can be observed from Table~\ref{table2} that our WDER-OC achieves the highest average returns in three out of four Mujoco tasks and is competitive in the remaining task, which demonstrates the improvement of WDEC-OC over the original OC regarding the performance and the sample efficiency.

Finally, we demonstrate the effectiveness of the WDER method by enabling agents to learn complex behaviors. Specifically, we showcase the contrasting performances of the baseline DEOC and the WDER method in Ant, Halfcheetah, Hopper, and Walker environments. Despite being the second best-performing algorithm in average return among the four tasks, the DEOC method still encounters scenarios that frequently lead to task failures in each environment. The scenarios are manifested as spinning in place (first four images in the top row of Fig.~\ref{figure5}), the sinking of the half-cheetah's head (first four images in the second row of Fig.~\ref{figure5}), stumbling and falling during landing due to an unstable center of gravity (first four images in the third row of Fig.~\ref{figure5}), and toppling over with excessive forward lean during leg swapping (first four images in the fourth row of Fig.~\ref{figure5}). In contrast, our WDER has learned diverse subpolicies that form more effective combination policies, significantly reducing the occurrences of task failures and achieving stable and outstanding performance.

\section{Conclusion}

This paper proposed a novel solution to the automated subpolicies discovery problem in HRL by introducing a task-agnostic regularizer, WDER, 
based on Wasserstein distance. Theoretically, the upper bound of the diversity of subpolicies generated by our approach is far larger than that of other algorithms utilizing information-theoretic objectives, and the gradients are more stable and effective throughout the updating process. We also demonstrated the effectiveness of our approach through extensive evaluations in two popular HRL task domains. The experimental results demonstrated that our method's robustness and generalization ability is higher than existing algorithms. Our future work will focus on an efficient Wasserstein distance estimation method to deal with situations involving more subpolicies.


\subsubsection{Acknowledgements} This work was supported in part by the National Key Research and Development Program of China under Grant 2020AAA0103405, the National
Natural Science Foundation of China under Grants 72293573 and 72293575, as well as the
Strategic Priority Research Program of Chinese Academy of Sciences under
Grant XDA27030100. 

\bibliography{ref_v6}
\end{document}